# MULTI MODEL DATA MINING APPROACH FOR HEART FAILURE PREDICTION

Priyanka H U[1] and Vivek R[2]

[1,2] Department of Information Science, RVCE, Bengaluru, India

*ABSTRACT*

*Developing predictive modelling solutions for risk estimation is extremely challenging in health-care informatics. Risk estimation involves integration of heterogeneous clinical sources having different representation from different health-care provider making the task increasingly complex. Such sources are typically voluminous, diverse, and significantly change over the time. Therefore, distributed and parallel computing tools collectively termed big data tools are in need which can synthesize and assist the physician to make right clinical decisions. In this work we propose multi-model predictive architecture, a novel approach for combining the predictive ability of multiple models for better prediction accuracy. We demonstrate the effectiveness and efficiency of the proposed work on data from Framingham Heart study. Results show that the proposed multi-model predictive architecture is able to provide better accuracy than best model approach. By modelling the error of predictive models we are able to choose sub set of models which yields accurate results. More information was modelled into system by multi-level mining which has resulted in enhanced predictive accuracy.*

## 1. INTRODUCTION

Present clinical and pharmaceutical environment are data intensive. Large amounts of data such as patient's narratives, scan reports, clinical, laboratory tests, and hospital administrative data are being produced routinely [1].This clinical data is digitized as computer records than maintaining it in the form of physical files. This diversified information may be from different systems and geographically separated too. 80% of clinically relevant data is unstructured [2]. This data is stored in multiple repositories as individual EMRs (Electronic Medical Records) in clinical laboratories and scanning image systems, they are also maintained as case files which consists of physician notes, medical correspondence, claims, CRM (Clinical Record Management) systems and finance. Patient care can be greatly improved by improving the means to access this valuable data, by improving the methodologies to parse and filtering this data into clinical and advanced analytics.

The main motivation behind digitization of health records is to lower the cost of health-care and reduce the number of preventable errors. But sheer amount of data collected, poses new challenges. Physicians are often overloaded from information gathered by various tools displaying large and irrelevant information. The issue is related to presenting the relevant information to the physician. The challenge lies in making patient data easily searchable and accessible, synthesize and assist the physician to make right clinical decisions. The challenges in working with clinical data are to integrate heterogeneous, multi standard data and standardize the meaning and representation of the integrated data.

Looking into the insight of big data, health-care falls into this category because of the large and complex data that is generated every single day. This data is difficult to manage by the traditional software and the hardware. By gaining knowledge in form insights provided by big data analytics significant improvement can be brought at lower costs. Physicians and doctors will be able to make better informed decision by exploiting explosion in data to extract insights [2] and thus making big data analytic application a current and trending research topic to work for.

The data intensive applications need distributed processing model. Apache Hadoop has a map reduce framework that processes the data of such applications. Azure HDInsight provides a software framework through which one can manage, analyse and report on big data. It deploys and provisions Apache Hadoop clusters in the cloud infrastructure. The Hadoop core provides a software framework to build applications that utilize the Hadoop Distributed File System (HDFS), and a simple Map-reduce programming model to process and analyse, in parallel, the data stored in this distributed system. Today, one of the major causes of death in adults with age greater than 65 is heart failure [3]. In our experiment we observe that a 1500 patient data-set with 14 attributes and with the single model prediction accuracy was approximately 82%. Whereas, with 25 different predictive models processed parallel, on the same dataset, 86% accuracy was achieved. This demonstrates the effectiveness of our methodology.

The rest of the paper is divided in following sections: in contribution section we will explain the novelty of the proposed methodology, detailed explanation of the multi model approach is given in the section Proposed methodology, effectiveness of the proposed multi model approach is observed in Results and discussion section, related work present in the literature are discussed in Related work section.

## 2. RELATED WORK

A number of researches have been undertaken in the field of healthcare. There is a vast amount of data which is unused. This data is growing rapidly in terms of size, complexity and speed of generation [4]. This paper presents a review of various algorithms from 1994-2013 necessary for handling such large data set. Definition of the process of generating a context-based view of the patient's health record is mentioned in [5]. It can be summarized in three steps 1) aggregate the patient data from separate clinical sources; 2) structure the patient record to identify problems, findings, and attributes reported in clinical reports, and map them to available knowledge sources; and 3) generate a tailored display based on annotations provided by the knowledge sources. But their work, says that the disease model may not fully meet the information needs of a physician performing a specific task. All the data elements displayed may be irrelevant.

Data Mining and Decision Making becomes extremely inefficient because of the variability in the stored records or accuracy of an individual decision making algorithm. The efficiency of individual algorithm has non zero variability. There can be many ways combining the predictions of which averaging is one method .While combining the different predictions, output may not be better than the best algorithm used. Many reviews refer to Dasarathy and Sheela's 1979 work as one of the earliest example of combined decision systems [6], with their ideas decision making-based algorithms, extending the boosting concept to multiple class and regression problems [7]. On partitioning the features using multiple algorithms. About a decade later, Hansen and Salamon showed that decision system of similarly configured neural networks can be used to improve classification performance [8]. However, it was Schapire's work that demonstrated through a procedure he named boosting that a strong classifier with an arbitrarily low error on a binary classification problem, can be constructed from a set of decision making algorithms, the error of any of which is merely better than that of random guessing [9]. The theory of boosting provided the foundation for the subsequent suite of AdaBoost algorithms.

With all this research, we justify that heart failure prediction using big data analytics is one of the most interesting and challenging tasks. The shortage of domain specialists and high number of wrongly diagnosed cases has necessitated the need to develop a fast and efficient detection system. We realize the need of efficient combining strategy of predictions generated by different algorithms. The strengths of parallel programming and Map reduce strategy are utilized through this study.

## 3. PROPOSED METHODOLOGY

Most of the heart failure predictive systems [10, 11] follow best model approach, they either tune the model or transform the available data for the better performance of the predictive systems. Such systems involve complex mathematical models and are difficult to convert to optimized applications; they lack the generality where they can be easily converted to a full-fledged product. Data transformation may result in either addition of redundant information or loss of valuable information from the data. In the proposed multi model approach we try bring in the required generality, by using models in their absolute form. Open source data mining tools are used to generate such predictive models. We employ both classification and clustering techniques with an aim to model more information into the system. Classification models are built for prediction and error of these models are clustered which helps in deciding the participation of models in the prediction. The proposed methodology was implemented in two phase, phase 1: development of predictive models and consolidation of the predictions using static weights, phase 2: development of clustering based error model which would have dual purpose of deciding the participation of model in the prediction and calculation of the dynamic weight.

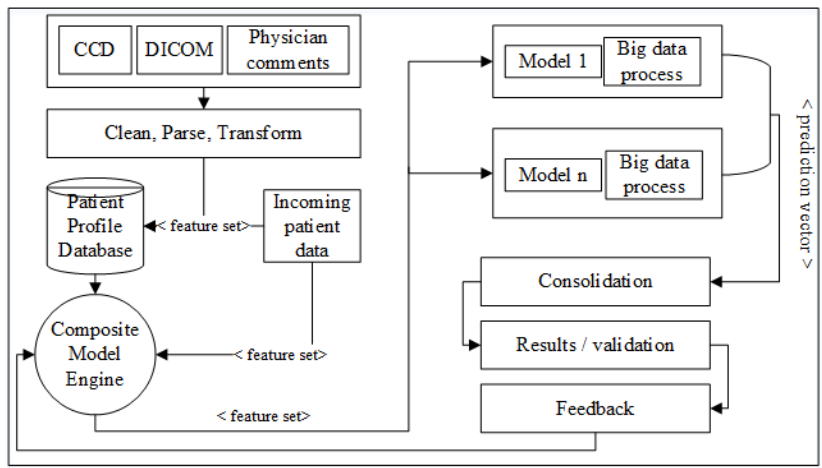

Figure 1: Proposed Architecture for multi model approach

Figure 1 shows the architecture of Multi model approach in its full capacity where in could be adopted as product. We briefly explain each component here. The patient profile database is the repository or the warehouse consisting of medical history of patients, many standards like CCD (continuity of care document) [16], DICOM [17], have evolved which efficiently represents the patients' medical condition over a period of time. These records are cleaned, parsed and transformed to suite mining activity before storing to warehouse. In our implementation we assume required dimensions are already extracted. The modules are required for cleaning, parse transform are not implemented.

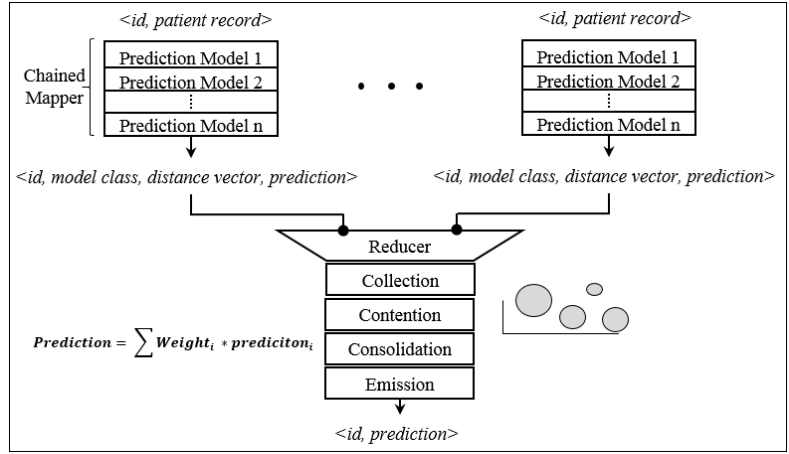

Figure 2: Program paradigm

Figure 2 shows the programing paradigm followed to implement the multi model architecture. Each trained model is converted to model class, providing interface for prediction; the mapper job invokes the prediction methods of all models configured, emitting a prediction vector $\rho_i$ <*id, model class, distance vector, prediction*> for each model, patient id and model class identifies for the key. The consolidation engine is implemented as reducer job, which collects, performs contention, consolidates and emits the final prediction. Each step is explained in detail below.

- Collection: A dictionary of prediction is created with patient id as key and list of prediction as value, whenever the reducer job comes across a prediction it be will added to list of prediction corresponding to patient id. Contention for participating model will start once list of prediction contains prediction vector from all configured model.

- Contention: In this step the participating model will be chosen out of configured model. This is an enhancement in phase two, as compared to phase one where all models would participate in final prediction with static weight of impact on final predication. Contention resolution is done by error modelling. Prediction value generated by models on training dataset can be clustered into 4 group clusters as shown in Table 2. Distance of current patient record from these four cluster is used as contention resolution criteria. A subset of configured models is chosen as participating models which are most likely to predict accurately.

- Consolidation: In this step normalized inverse distance function of participating models are calculated, which forms the dynamic weight of impact, equation 5 is used to get the final prediction.

- Emission: patient id and prediction value is emitted by reducer and also deletes the dictionary entry for corresponding patient.

As explained above, Multi model prediction approach was implemented in two phase. In phase one the impact of participating model in final prediction will be fixed, i.e., the predictive weight assigned to each model in final prediction is simply $1/n$, where n is the number of predictive models used. The final prediction with static weights (ρm) is given by equation 1, where ρi is the prediction made by individual prediction model.

$$\rho_m = \sum_{i=1}^{\#models} \rho_i * \frac{1}{n} \qquad (1)$$

International Journal of Data Mining & Knowledge Management Process (IJDKP) Vol.6, No.5, September 2016In this approach the effect of poorly performing models will nullify the effect of correctly predicting models. The aim of phase two implementation is to provide a model filtering process which will select only subset of better of performing model and yield to better predictive accuracy.

Table 1. Cluster categories used in error model

| Cluster | Description |
|---|---|
| Cluster 00 *(C00)* | cluster of records where in the prediction class is 0 and model predicted class is also 0 *(true Negative prediction)* |
| Cluster 01 *(C01)* | cluster of records where in the prediction Class is 0, but predicted class is 1 *(false positive prediction)* |
| Cluster 10 *(C10)* | Cluster of records where in the prediction class is 1, but the predicted class is 0 *(false negative prediction)* |
| Cluster 11 *(C11)* | Cluster of records where in the prediction class is 1 and model predicted class is also 1 *(true positive prediction)* |

The erroneous predictions made by models are used as feedback to determine the most suitable subset of participating models, which are likely to predict accurately. For each predicting model the training set is divided into four error clusters C00, C01, C10, C11, Table 1 gives explanation of each cluster. Under testing, each tuple $t_i$ whose class is to predicted, distance of $t_i$ from each of the error clusters is calculated which gives us a distance vector < d00, d01, d10, d11 >, which determines the participation of given model in final prediction and also the impact (weight) the model's prediction carries in the final prediction. Listing 1 explains the logic which determines the participation of model.

Listing 1. Logic to determine participation of model in final prediction

*if $\rho_i \approx 0$ then,*
    *if $d_{00} < d_{01}$ then $i^{th}$ model will participate in final prediction*
*if $\rho_i \approx 1$ then,*
    *if $d_{11} < d_{10}$ then $i^{th}$ model will participate in final prediction*

Once we have a subset of participating models and their predicted class along with the distance vector Vi, for each class, we employ inverse distance function [17] given by equation 2 to determine the weight of participating model ($wt.d_i^{-1}$), these weights are then normalized between values 0 and 1 as given by equation 3, which forms the dynamic weight used for final prediction. Final prediction is given by equation 4. Where $\rho_{final}$ is the final prediction made by proposed multi model approach. One of the clear enhancement to proposed model is increasing the number of clusters in each cluster category i.e., clusters C00, C01, C10, C11 will represent set of clusters rather than single cluster, doing so will greatly enhance the efficiency of model to determine the participating models and their dynamic weights. This work is considered under future scope.

$$wt.d_i^{-1} = \frac{d_i}{\sum_{i=1}^{\#participating\,model} d_i} \quad (2)$$

$$normalized(wt.d_i^{-1}) = \frac{wt.d_i^{-1}}{\sum_{i=1}^{\#participating\,model} wt.d_i^{-1}} \quad (3)$$

$$\rho_{final} = \sum_{i=1}^{\#participating\,model} wt.d_i^{-1} * \rho_i \quad (4)$$

## 4. RESULTS AND DISCUSSION

Effectiveness of Multi model prediction was tested on the Framingham [14] dataset consisting of 1500 patient records. The same is a subset of data collected from the Framingham study and includes lab/clinical data of 1500 patients. Table 2 shows a snapshot of the Framingham dataset, the main aim of the multi model is to predict the class field in the data, i.e., to determine whether the patient will be alive or dead. More explanation of various fields can be found in [14].

Table 2. Framingham data snapshot

| Id | Sex | Age | FRW | SBP | DBP | CHOL | CIG | CHD | Class |
|---|---|---|---|---|---|---|---|---|---|
| 4988 | female | 57 | 135 | 186 | 120 | 150 | 0 | 1 | Alive |
| 3001 | female | 60 | 123 | 165 | 100 | 167 | 25 | 0 | Death |

For all model creation, Weka, a data mining tool was used [15], with 75% as training data and 25% as test data. The proposed multi model approach was implemented as Map Reduce streaming jobs suitable for running on Hadoop framework [16] under C# .Net framework. The proposed multi model approach is compared with best model approach. 25 prediction models were trained on the dataset, with best model being Regression by discretization providing predictive efficiency of 82.42%. Model efficiency is given by equation 1. Table 3 depicts various predictive models trained using Weka tool along with their predictive efficacy.

$$Effciency = \frac{\#Patient\ Class\ correctly\ predicted}{\#Total\ number\ of\ Patients\ in\ dataset} \quad (5)$$

Table 3. Different predictive models trained using Weka tool

| Weka Model Name | Efficiency |
|---|---|
| Bagging (A to I) | 80.91 |
| Random Sub space (A to J) | 79.62 |
| M5Prune Model | 79.12 |
| Linear Regression | 77.33 |
| Pace Regression | 76.97 |
| Regression By Discretization | 82.42 |
| REP tree | 77.90 |

As explained in proposed methodology section, Multi model prediction approach was implemented in two phase. In phase one where the each predictive models was given a static weight of 0.04 (which is 1/25 models), under such setup the multi model predictor's efficiency was observed to be 84.36%. In phase two implementation, where a subset of model was chosen using the process explained in proposed methodology, under such a setup the multi model predictors efficiency was observed to be 85.87%. Table 4 consolidates and compares the efficiency of proposed model, from the observed results we can say that proposed multi model approach out performs the best model approach.

Table 4. Consolidated Result

| Approach | Efficiency |
|---|---|
| Best Model | 82.42% |
| Multi Model with all models participating and static weights. | 84.36% |
| Multi model with best models participating and dynamic weights | 85.87% |

## 5. CONCLUSIONS

In this paper we analysed a generic architecture for clinical decision making and predictive modelling system that feeds on already existing information management system or enterprise healthcare data warehouse. Results show that the proposed multi model predictive architecture is able to provide better accuracy than best model approach. By modelling the error of predictive models we are able choose sub set of models which yields accurate results. We were able to model more information into system by multi-level mining which has resulted in enhanced predictive accuracy.